\documentclass{article}

\usepackage{arxiv}
\date{}
\renewcommand{\today}{}

\usepackage[utf8]{inputenc} 
\usepackage[T1]{fontenc}    
\usepackage{hyperref}       
\usepackage{url}            
\usepackage{booktabs}       
\usepackage{amsfonts}       
\usepackage{nicefrac}       
\usepackage{microtype}      
\usepackage{amsmath}        
\usepackage{lipsum}
\usepackage{graphicx}
\graphicspath{ {./images/} }
\usepackage{float}
\usepackage{natbib}
\usepackage{xcolor}
\usepackage{wrapfig}
\definecolor{Highlight}{RGB}{0,150,0}

\usepackage{algorithm}
\usepackage{algpseudocode}   
\usepackage{caption}   
\usepackage[table]{xcolor}
\usepackage{graphicx}
\usepackage{booktabs}
\usepackage{multirow}
\usepackage{amsmath,amssymb}
\usepackage{algorithm}
\usepackage{algpseudocode}
\usepackage{tabularx}
\setcitestyle{numbers}

\definecolor{hotpurple}{RGB}{180, 80, 200}

\title{Latent Transfer Attack: Adversarial Examples via Generative Latent Spaces}

\newcommand{\ourmethod}{\textsc{LTA}}

\author{
  Eitan Shaar$^{1}$\thanks{Equal contribution, Corresponding author: shaarei@biu.ac.il, arielshaulov@mail.tau.ac.il} \quad
  Ariel Shaulov$^{2*}$ \quad
  Yalcin Tur$^3$ \quad
  Gal Chechik$^{4,5}$ \quad
  Ravid Shwartz-Ziv$^6$ \\ \\
$^{1}$Independent Researcher \quad
$^{2}$Tel-Aviv University \quad
$^{3}$Stanford University \quad \\
$^{4}$Bar Ilan University  \quad
$^{5}$NVIDIA Research \quad
$^{6}$New York University \quad
}

\begin{document}
\maketitle

\begin{abstract}
Adversarial attacks are a central tool for probing the robustness of modern vision models, yet most methods optimize perturbations directly in pixel space under $\ell_\infty$ or $\ell_2$ constraints.
While effective in white-box settings, pixel-space optimization often produces high-frequency, texture-like noise that is brittle to common preprocessing (e.g., resizing and cropping) and transfers poorly across architectures.
We propose \textbf{LTA} (\textbf{L}atent \textbf{T}ransfer \textbf{A}ttack), a transfer-based attack that instead optimizes perturbations in the latent space of a pretrained Stable Diffusion VAE. Given a clean image, we encode it into a latent code and optimize the latent representation to maximize a surrogate classifier loss, while softly enforcing a pixel-space $\ell_\infty$ budget after decoding. To improve robustness to resolution mismatch and standard input pipelines, we incorporate Expectation Over Transformations (EOT) via randomized resizing, interpolation, and cropping, and apply periodic latent Gaussian smoothing to suppress emerging artifacts and stabilize optimization.
Across a suite of CNN and vision-transformer targets, LTA achieves strong transfer attack success while producing spatially coherent, predominantly low-frequency perturbations that differ qualitatively from pixel-space baselines and occupy a distinct point in the transfer–quality trade-off. Our results highlight pretrained generative latent spaces as an effective and structured domain for adversarial optimization, bridging robustness evaluation with modern generative priors.
\end{abstract}

\section{Introduction}
\label{sec:intro}

Transferability is a fundamental bottleneck for black-box adversarial attacks.
Most existing methods optimize perturbations directly in pixel space under $\ell_\infty$ or $\ell_2$ constraints. While highly effective in white-box settings, pixel-space gradients naturally exploit high-frequency, non-robust features that are architecture-specific~\cite{ilyas2019adversarial,tsipras2019robustness}: the resulting perturbations appear as texture-like noise, are brittle to standard preprocessing such as resizing and cropping~\cite{engstrom2019rotation,xie2019improving}, and transfer poorly across architectures (e.g., from CNNs to vision transformers)~\cite{dong2018boosting,papernot2017transferability}. This suggests that pixel space may be a suboptimal domain for constructing perturbations that are simultaneously effective, transferable, and visually coherent, and that constraining perturbations to lower-frequency, more structured variations may improve cross-model transfer~\cite{engstrom2019rotation,naseer2021generating}.



Recent advances in large-scale generative modeling offer a principled alternative. Variational Autoencoders (VAEs) trained on diverse natural images learn compact latent representations that capture semantic and structural regularities of the data distribution~\cite{kingma2014autoencoding,rezende2014stochastic}. Crucially, the decoder of such a VAE imposes a strong inductive bias: small perturbations in latent space decode into spatially smooth, predominantly low-frequency variations in pixel space. We argue that this frequency bias is precisely what enables improved adversarial transferability, perturbations that are structurally aligned with the image manifold are more likely to exploit features shared across architectures and to survive standard input pipelines. In particular, the VAE component of Stable Diffusion, trained on a large and diverse image corpus, provides a broadly useful image prior whose latent space is well-suited to this purpose.


In this work, we introduce \textbf{\ourmethod} (Latent Transfer Attack), a transfer-based attack that generates adversarial examples by optimizing perturbations in the latent space of a pretrained Stable Diffusion VAE rather than directly in pixel space. By leveraging the VAE decoder as an implicit image prior, our method naturally restricts perturbations to structured, low-frequency directions aligned with the image manifold. A practical challenge arises from the mismatch between the resolution and preprocessing used by the generative model and the input pipelines of downstream classifiers. Perturbations optimized on a fixed decoded image may not survive common transformations such as resizing, interpolation, or cropping, which can significantly reduce transferability.  To address this issue, we optimize the attack under Expectation Over Transformations (EOT)~\cite{athalye2018obfuscated}, sampling random resizing, interpolation, and cropping operations during optimization to explicitly model these pipelines. In addition, direct optimization in latent space can accumulate localized artifacts that destabilize the optimization trajectory. 
We therefore introduce a lightweight periodic smoothing step in latent space that suppresses emerging high-frequency components while preserving the global structure of the perturbation.

These components support the central idea of latent-space adversarial optimization rather than constituting independent contributions. Extensive experiments demonstrate that \ourmethod{} substantially improves transfer attack success across a wide range of target architectures, with particularly strong gains in CNN-to-ViT transfer and against purification-based defenses. We additionally provide a frequency-domain analysis showing that latent-space perturbations concentrate energy in low-frequency bands, offering insight into why generative latent spaces constitute an effective domain for adversarial optimization.


Our main contributions are:
\begin{itemize}


\item We propose \ourmethod, a simple and effective framework that performs adversarial optimization in the latent space of a pretrained generative VAE, leveraging the decoder as an implicit low-frequency image prior to improve cross-architecture transfer.

\item We provide a frequency-domain analysis demonstrating that latent-space optimization naturally biases adversarial perturbations toward low-frequency components, and connect this spectral property to the observed gains in cross-architecture and cross-defense transfer.


\item \ourmethod{} achieves state-of-the-art transferability across a diverse suite of CNN and ViT targets, with the largest gains in the challenging CNN$\to$ViT setting ($+13.7$ points averaged over transformer targets with RN50 surrogate) and under purification-based defenses (up to $+34.3$ points).

\end{itemize}

\section{Related Work}
\label{sec:related_work}

\subsection{Pixel-Space Adversarial Attacks}

Early work on adversarial examples demonstrated that deep neural networks are highly vulnerable to small, carefully constructed input perturbations~\cite{szegedy2014intriguing}. Subsequent gradient-based methods such as FGSM~\cite{goodfellow2015explaining}, BIM~\cite{kurakin2017adversarial}, PGD~\cite{madry2018towards}, and C\&W~\cite{carlini2017towards} established standard benchmarks for robustness evaluation under $\ell_\infty$ and $\ell_2$ constraints, while further extensions explored stronger iterative strategies and confidence-based objectives~\cite{brendel2018decisionbased,moosavi2016deepfool}.


However, pixel-space attacks typically exploit non-robust, high-frequency features that are weakly aligned with human perception~\cite{ilyas2019adversarial,tsipras2019robustness}. Several works have shown that these perturbations correspond to fragile directions in input space rather than semantically meaningful changes~\cite{fawzi2018analysis,tanay2016boundary}, making them sensitive to common preprocessing operations such as resizing, cropping, and interpolation~\cite{engstrom2019rotation,guo2018countering,xie2019improving}. Moreover, pixel-space attacks exhibit limited transferability across architectures with different inductive biases, particularly from CNNs to vision transformers~\cite{dong2018boosting,papernot2016transferability}. These limitations have motivated research into alternative perturbation domains that yield more structured and transferable adversarial signals~\cite{sharma2019beyond,wu2020adversarial}.



\subsection{Transfer-Based Adversarial Attacks}

To address the limited generalization of pixel-space attacks, a large body of work has focused on improving transferability in black-box settings, where adversarial examples are generated on a surrogate model and evaluated on unseen targets.
Early approaches introduced momentum~\cite{dong2018boosting}, input diversity~\cite{xie2019improving}, and translation invariance~\cite{dong2019evading} to reduce overfitting to the surrogate and encourage perturbations that generalize across models.
Related works further explored random transformations and expectation over transformations (EOT) to improve robustness against preprocessing and defenses~\cite{athalye2018obfuscated,guo2018countering}.

More recent methods seek to explicitly manipulate intermediate representations rather than raw pixels.
Feature-level and representation-alignment attacks encourage perturbations that disrupt shared internal features across architectures~\cite{naseer2021generating,li2023improving,huang2021intermediate}.
Other approaches leverage ensembles of surrogate models or optimize perturbations jointly across multiple networks to improve cross-model generalization~\cite{tramer2017ensemble,liu2017delving}.
Frequency-aware and spectral attacks have also been proposed to exploit low-frequency components that are more transferable across models~\cite{gupta2023semantic,sharma2019highfrequency,tsipras2019robustness}.

State-of-the-art transfer attacks such as BFA~\cite{wang2024improving}, P2FA~\cite{liu2025pixel2feature}, and MFAA~\cite{zheng2025enhancing} further refine this idea by exploiting feature hierarchies, attention mechanisms, or frequency-aware objectives~\cite{wang2024boostingtransferabilityadversarialattacks,fang2024strong,zheng2025enhancing}.
Despite these advances, most transfer-based attacks remain fundamentally tied to pixel-space gradients from discriminative classifiers. Our approach instead performs adversarial optimization in the latent space of a pretrained generative model, providing a structured prior that naturally promotes cross-architecture transfer.



\subsection{Generative Models for Adversarial Example Generation}

Generative models have recently been explored as a means of imposing structure and semantic constraints on adversarial perturbations.
Early work employed GANs to directly generate adversarial examples in a feed-forward manner~\cite{advgan,poursaeed2018generative}, enabling fast attacks after training.
Other approaches constrained perturbations to lie on learned image manifolds, producing so-called semantic or manifold-based adversarial examples that modify higher-level attributes rather than individual pixels~\cite{rozsa2016adversarial,song2018constructing,gupta2023semantic}.

Autoencoder and VAE-based methods have also been proposed to regularize perturbations through reconstruction losses or latent constraints, aiming to suppress visually implausible noise~\cite{schott2019towards,joshi2020towards}.
These approaches typically operate by projecting perturbations onto a learned data manifold or optimizing adversarial directions in a latent space rather than in the input domain.

More recently, diffusion models have emerged as powerful priors for adversarial generation.
DiffAttack and related methods exploit diffusion processes to synthesize adversarial images that are more natural-looking and robust to defenses~\cite{chen2024diffusion,nie2022diffusion,huang2023advdiffusion}. These approaches typically rely on iterative denoising procedures in pixel space or train auxiliary generative networks for attack generation.

Our method differs fundamentally from prior generative attacks in that it directly optimizes the latent representation of a pretrained diffusion VAE rather than training a new generator or performing pixel-space diffusion.
By leveraging the decoder as an implicit image prior, latent-space optimization restricts perturbations to spatially coherent, low-frequency directions aligned with the natural image distribution. This enables strong transferability without requiring additional generative training or full diffusion sampling, and provides a simple yet effective bridge between adversarial optimization and modern generative latent spaces.

\section{Method}
\label{sec:method}

We propose \textbf{\ourmethod}, a transfer-based attack that generates adversarial examples by optimizing perturbations in the latent space of a pretrained Variational Autoencoder (the Stable Diffusion VAE), rather than directly in pixel space.
The key idea is to leverage the VAE decoder as a differentiable mapping from a compact latent space to pixel space, so that adversarial optimization is implicitly constrained by the structure of the learned representation.
Below we describe the setup, objective, and two supporting components: expectation over transformations and periodic latent smoothing. The full procedure is summarized in Algorithm~\ref{alg:ourmethod}.


\subsection{Setup}

Let $f(\cdot)$ be a surrogate classifier and $\ell(\cdot,y)$ a classification loss for label $y$.
Let $\mathrm{Enc}$ and $\mathrm{Dec}$ denote the frozen VAE encoder and decoder.
Given an image $x\in[0,1]^{H\times W\times 3}$, we encode it to a latent code $z_0=\mathrm{Enc}(x)$ and optimize a latent variable $z$.
The adversarial image is obtained by decoding: $x_{\text{adv}}=\mathrm{Dec}(z)$.

\subsection{Objective}
We maximize the surrogate classification loss under randomized preprocessing while softly enforcing a pixel-space $\ell_\infty$ budget after decoding.
Specifically, we use the standard cross-entropy loss $\ell_{\mathrm{CE}}$ and minimize

\begin{equation}
\mathcal{L}(z)
=
-\underbrace{\mathbb{E}_{t\sim\mathcal{T}}
\big[\ell_{\mathrm{CE}}\big(f(t(\mathrm{Dec}(z))),y\big)\big]}_{\mathcal{L}_{\text{EOT}}(z)}
\;+\;
\lambda_\epsilon
\underbrace{\frac{1}{|x|}\sum_i \mathrm{ReLU}\!\big(|x_{\text{adv},i}-x_i|-\epsilon\big)}_{\mathcal{L}_\epsilon(z)} ,
\label{eq:overall_obj}
\end{equation}

where $\mathcal{T}$ is a distribution over input transformations, $\epsilon$ is the target $\ell_\infty$ budget in pixel space, and $\lambda_\epsilon$ controls the penalty strength. The soft penalty $\mathcal{L}_\epsilon$ is zero within the budget and increases only for violations. We use a soft penalty rather than hard projection because the mapping from latent space to pixel space is nonlinear: clipping in pixel space and re-encoding would not preserve the latent structure, and clipping directly in latent space has no principled correspondence to a pixel-space $\ell_\infty$ constraint. We optimize $z$ using Adam, which we find more effective than SGD with momentum for navigating the curved latent geometry.


\subsection{Expectation Over Transformations}
\label{sec:eot}

Unlike pixel-space attacks that can directly produce outputs at the target classifier's resolution, the VAE decoder outputs at a fixed resolution (e.g., $256{\times}256$) that generally differs from standard classifier inputs (e.g., $224{\times}224$). This resolution mismatch, combined with sensitivity to interpolation and cropping, motivates the use of expectation over transformations (EOT):
at each iteration, we sample $K$ transforms $\{t_k\}_{k=1}^K\sim\mathcal{T}$ and average the loss,


\begin{equation}
\mathcal{L}_{\text{EOT}}(z)\approx \frac{1}{K}\sum_{k=1}^K \ell\big(f(t_k(\mathrm{Dec}(z))),y\big).
\label{eq:eot}
\end{equation}
We instantiate $\mathcal{T}$ with random resize, random interpolation kernel, and near-center crop with jitter. This encourages perturbations that are effective across a range of preprocessing pipelines rather than overfitting to a single resize-and-crop configuration.

\subsection{Periodic Latent Smoothing}

Although the VAE decoder biases perturbations toward low-frequency structure, iterative optimization over many steps can still accumulate localized, high-frequency artifacts in the latent code. To counteract this, we regularize the latent perturbation $\Delta z=z-z_0$ by periodically smoothing it:

\begin{equation}
\Delta z \leftarrow G * \Delta z,
\qquad
z \leftarrow z_0+\Delta z,
\label{eq:smooth}
\end{equation}
where $G$ denotes a small Gaussian smoothing kernel applied via depthwise convolution. This operation is applied every $N$ steps and acts as a lightweight regularizer that suppresses emerging artifacts without significantly constraining the adversarial signal.

\begin{algorithm}[t]
\caption{\ourmethod: VAE-latent adversarial optimization}
\label{alg:ourmethod}
\begin{algorithmic}[1]
\Require Clean image $x$, label $y$, encoder/decoder $(\mathrm{Enc},\mathrm{Dec})$, surrogate $f$, budget $\epsilon$
\Require Steps $T$, EOT samples $K$, smoothing period $N$, penalty weight $\lambda_\epsilon$, Adam params $(\alpha,\beta_1,\beta_2)$
\State $z_0 \gets \mathrm{Enc}(x)$; \quad $z \gets z_0$
\For{$t = 1$ \textbf{to} $T$}
    \State $x_{\text{adv}} \gets \mathrm{Dec}(z)$
    \State Sample $\{t_k\}_{k=1}^{K} \sim \mathcal{T}$
    \State $\mathcal{L}_{\text{EOT}} \gets \frac{1}{K}\sum_{k=1}^{K} \ell\!\big(f(t_k(x_{\text{adv}})), y\big)$
    \State $\mathcal{L}_{\epsilon} \gets \frac{1}{|x|}\sum_{i}\mathrm{ReLU}\!\big(|x_{\text{adv},i}-x_i|-\epsilon\big)$
    \State $\mathcal{L} \gets -\mathcal{L}_{\text{EOT}} + \lambda_{\epsilon}\mathcal{L}_{\epsilon}$ \Comment{Eq.~\eqref{eq:overall_obj}}
    \State Update $z$ with Adam to minimize $\mathcal{L}$
    \If{$t \bmod N = 0$}
        \State $\Delta z \gets z - z_0$
        \State $\Delta z \gets G_{\sigma} * \Delta z$
        \State $z \gets z_0 + \Delta z$
    \EndIf
\EndFor
\State \Return $x_{\text{adv}} \gets \mathrm{Dec}(z)$
\end{algorithmic}
\end{algorithm}

\section{Experiments}
\subsection{Setup}
\label{sec:exp_setup}

\paragraph{\textbf{Dataset.}}
We adopt an ImageNet-compatible evaluation set comprising 1,000 PNG images sampled from the ImageNet validation distribution using a fixed random seed.

\paragraph{\textbf{Threat model and evaluation.}}

We consider a standard transfer-based black-box setting: adversarial examples are optimized on a surrogate classifier and then evaluated on a set of unseen target models.
We report Attack Success Rate (ASR\%; higher is better for the attacker) on each target, as well as defense-robust ASR when attacks are evaluated through defenses.

\paragraph{\textbf{Surrogate and target models.}}

We use three surrogate networks: ResNet-50 (RN50), ResNet-152 (RN152), and VGG-16 (VGG16)~\cite{he2016deep,simonyan2014very}.
Transferability is evaluated on the following targets:
RN50, RN152, VGG16, MobileNetV2 (MNv2)~\cite{sandler2018mobilenetv2},
Inception-V3 (IncV3)~\cite{szegedy2016rethinking},
ViT-B/16~\cite{dosovitskiy2020image}, PiT-B~\cite{heo2021rethinking},
Visformer-S~\cite{chen2021visformer}, and Swin-T~\cite{liu2021swin}.
All models use standard ImageNet preprocessing.

\paragraph{\textbf{Defense models.}}
To evaluate robustness beyond standard (undefended) classifiers, we additionally test transfer attacks under five widely-used defense pipelines:

(i)~\textbf{AT}, adversarially trained models, which improve robustness by training on adversarial examples (we use the commonly adopted ensemble/adv-trained setting)~\cite{tramer2017ensemble};
(ii)~\textbf{HGD}, High-level Representation Guided Denoiser, which purifies adversarial inputs by training a denoiser with a loss defined in the target model's feature/logit space~\cite{liao2018defense};
(iii)~\textbf{RS}, Randomized Smoothing, which certifies/predicts via Gaussian-noise aggregation around the input~\cite{cohen2019certified};
(iv)~\textbf{NRP}, Neural Representation Purifier, a self-supervised purification module trained to align deep representations of perturbed images with clean ones~\cite{naseer2020self};
and (v)~\textbf{DiffPure}, diffusion-based purification that removes adversarial perturbations via a short forward diffusion step followed by reverse denoising~\cite{nie2022diffusion}.

\paragraph{\textbf{Baselines.}}
We compare against strong recent transfer attacks spanning feature-level, gradient-based, and diffusion-based paradigms:
P2FA~\cite{liu2025pixel2feature}, BFA~\cite{wang2024improving}, MFAA~\cite{zheng2025enhancing},
ANDA~\cite{fang2024strong}, GI-FGSM~\cite{wang2024boostingtransferabilityadversarialattacks},
ILPD~\cite{li2023improving}, and the diffusion-based DiffAttack~\cite{chen2024diffusion}.

\paragraph{\textbf{Implementation details.}}
We fix a single random seed for all runs. \ourmethod{} uses $\epsilon{=}16/255$ (pixel-space after decoding) and optimizes VAE latents with Adam (lr $0.1$) for $T{=}150$ steps. We use EOT with $K{=}4$ transforms per step (random resize in $[224,288]$, random interpolation, and near-center crop with up to $8$px jitter), a soft pixel penalty weight $\lambda_\epsilon{=}10.0$, and latent Gaussian smoothing every $N{=}10$ steps with a $3{\times}3$ kernel ($\sigma{=}0.5$). Encoding/decoding is performed at $256{\times}256$ using the Stable Diffusion v1.5 VAE. See hyperparameter analysis in Supp~\ref{sec:hparam_analysis}. For baselines, we use their official/public implementations with recommended hyperparameters, while matching the dataset, preprocessing, and evaluation pipeline across methods.

\begin{table*}[t]
\centering
\caption{\textbf{\ourmethod{} achieves the strongest transfer across all surrogates, with the largest gains on vision transformer targets.} Attack Success Rate (ASR\%, higher is better) across target models, grouped by surrogate.
Gray columns denote transformer targets. 
\textit{--}~indicates the target was not evaluated.}
\label{tab:asr_per_model}

\scriptsize
\renewcommand{\arraystretch}{1.08}

\resizebox{\textwidth}{!}{%
\begin{tabular}{ll ccccc cccc c}
\toprule
& & \multicolumn{5}{c}{\textbf{CNN Targets}} & \multicolumn{4}{c}{\textbf{ViT Targets}} & \\
\cmidrule(lr){3-7} \cmidrule(lr){8-11}
\textbf{Surrogate} & \textbf{Method} & \textbf{RN50} & \textbf{RN152} & \textbf{VGG16} & \textbf{MNv2} & \textbf{IncV3} & \cellcolor{gray!8}\textbf{ViT-B/16} & \cellcolor{gray!8}\textbf{PiT-B} & \cellcolor{gray!8}\textbf{Visf.-S} & \cellcolor{gray!8}\textbf{Swin-T} & \textbf{Avg.} \\
\midrule
\multirow{8}{*}{ResNet-50} 
 & P2FA & \textbf{100.0} & \textbf{98.8} & \textbf{98.2} & \textbf{97.8} & 83.1 & \cellcolor{gray!8}39.4 & \cellcolor{gray!8}65.5 & \cellcolor{gray!8}85.1 & \cellcolor{gray!8}82.5 & 83.4 \\
 & BFA & 98.1 & 94.1 & 93.7 & 92.8 & \textbf{85.7} & \cellcolor{gray!8}49.5 & \cellcolor{gray!8}72.1 & \cellcolor{gray!8}86.1 & \cellcolor{gray!8}83.4 & 82.7 \\
 & MFAA & 95.3 & 83.6 & 86.4 & 84.7 & 70.8 & \cellcolor{gray!8}24.0 & \cellcolor{gray!8}41.0 & \cellcolor{gray!8}59.8 & \cellcolor{gray!8}57.0 & 67.0 \\
 & ANDA & 99.9 & 84.6 & 83.9 & 82.8 & 73.4 & \cellcolor{gray!8}42.4 & \cellcolor{gray!8}61.6 & \cellcolor{gray!8}72.0 & \cellcolor{gray!8}66.6 & 74.1 \\
 & GI-FGSM & \textbf{100.0} & 65.0 & 72.0 & 65.3 & 49.2 & \cellcolor{gray!8}15.7 & \cellcolor{gray!8}28.1 & \cellcolor{gray!8}38.3 & \cellcolor{gray!8}36.2 & 52.2 \\
 & ILPD & 97.8 & 82.4 & 88.2 & 86.0 & 71.1 & \cellcolor{gray!8}46.1 & \cellcolor{gray!8}61.7 & \cellcolor{gray!8}69.7 & \cellcolor{gray!8}66.5 & 74.4 \\
 & DiffAttack & 92.7 & 51.0 & 50.6 & 51.5 & 47.0 & \cellcolor{gray!8}29.4 & \cellcolor{gray!8}43.7 & \cellcolor{gray!8}46.6 & \cellcolor{gray!8}47.7 & 51.1 \\
 \cmidrule(lr){2-12}
 & \ourmethod{} (Ours) & \textbf{100.0} & 96.3 & 92.8 & 91.4 & 82.9 & \cellcolor{gray!8}\textbf{71.3} & \cellcolor{gray!8}\textbf{91.1} & \cellcolor{gray!8}\textbf{92.1} & \cellcolor{gray!8}\textbf{91.4} & \textbf{89.9} \\
\midrule
\multirow{8}{*}{ResNet-152} 
 & P2FA & \textbf{99.7} & 99.9 & \textbf{98.4} & \textbf{98.5} & \textbf{95.3} & \cellcolor{gray!8}57.7 & \cellcolor{gray!8}81.8 & \cellcolor{gray!8}94.1 & \cellcolor{gray!8}90.0 & 90.6 \\
 & BFA & 96.4 & 98.1 & 93.8 & 93.1 & 89.8 & \cellcolor{gray!8}66.6 & \cellcolor{gray!8}83.9 & \cellcolor{gray!8}91.4 & \cellcolor{gray!8}88.1 & 89.0 \\
 & MFAA & 83.4 & 93.9 & 81.7 & 77.6 & 71.0 & \cellcolor{gray!8}43.6 & \cellcolor{gray!8}60.0 & \cellcolor{gray!8}69.7 & \cellcolor{gray!8}66.4 & 71.9 \\
 & ANDA & 94.3 & 99.8 & 84.9 & 81.7 & 81.7 & \cellcolor{gray!8}50.5 & \cellcolor{gray!8}71.0 & \cellcolor{gray!8}80.7 & \cellcolor{gray!8}74.1 & 79.9 \\
 & GI-FGSM & 88.2 & 99.5 & 77.7 & 72.7 & 62.6 & \cellcolor{gray!8}27.7 & \cellcolor{gray!8}45.3 & \cellcolor{gray!8}55.7 & \cellcolor{gray!8}49.6 & 64.3 \\
 & ILPD & 90.0 & 96.7 & 87.3 & 85.8 & 76.2 & \cellcolor{gray!8}55.0 & \cellcolor{gray!8}67.9 & \cellcolor{gray!8}75.8 & \cellcolor{gray!8}71.5 & 78.5 \\
 & DiffAttack & 68.1 & 89.3 & 56.4 & 54.7 & 54.9 & \cellcolor{gray!8}43.1 & \cellcolor{gray!8}56.5 & \cellcolor{gray!8}57.9 & \cellcolor{gray!8}56.9 & 59.8 \\
 \cmidrule(lr){2-12}
 & \ourmethod{} (Ours) & 99.2 & \textbf{100.0} & 90.0 & 89.6 & 86.1 & \cellcolor{gray!8}\textbf{81.7} & \cellcolor{gray!8}\textbf{94.0} & \cellcolor{gray!8}\textbf{95.6} & \cellcolor{gray!8}\textbf{95.6} & \textbf{92.4} \\
\midrule
\multirow{8}{*}{VGG-16} 
 & P2FA & 86.6 & 70.5 & \textbf{100.0} & 96.7 & 69.8 & \cellcolor{gray!8}15.2 & \cellcolor{gray!8}35.6 & \cellcolor{gray!8}69.9 & \cellcolor{gray!8}63.4 & 67.5 \\
 & BFA & 92.7 & 84.3 & \textbf{100.0} & 97.4 & 88.6 & \cellcolor{gray!8}30.4 & \cellcolor{gray!8}58.8 & \cellcolor{gray!8}82.8 & \cellcolor{gray!8}74.8 & 78.9 \\
 & MFAA & 79.4 & 64.0 & \textbf{100.0} & 93.1 & 78.1 & \cellcolor{gray!8}22.0 & \cellcolor{gray!8}36.1 & \cellcolor{gray!8}61.6 & \cellcolor{gray!8}54.5 & 65.4 \\
 & ANDA & 80.1 & 65.4 & \textbf{100.0} & 92.0 & 80.6 & \cellcolor{gray!8}24.7 & \cellcolor{gray!8}45.9 & \cellcolor{gray!8}67.0 & \cellcolor{gray!8}63.0 & 68.7 \\
 & GI-FGSM & 66.8 & 47.3 & 99.9 & 87.1 & 63.4 & \cellcolor{gray!8}16.7 & \cellcolor{gray!8}31.7 & \cellcolor{gray!8}49.1 & \cellcolor{gray!8}49.1 & 56.8 \\
 & ILPD & 91.8 & 83.3 & 99.8 & 97.2 & 86.2 & \cellcolor{gray!8}32.9 & \cellcolor{gray!8}60.1 & \cellcolor{gray!8}83.2 & \cellcolor{gray!8}79.0 & 79.3 \\
 & DiffAttack & 71.5 & 57.7 & 97.1 & 78.8 & 68.5 & \cellcolor{gray!8}38.8 & \cellcolor{gray!8}54.4 & \cellcolor{gray!8}60.4 & \cellcolor{gray!8}64.2 & 65.7 \\
 \cmidrule(lr){2-12}
 & \ourmethod{} (Ours) & \textbf{99.8} & \textbf{99.0} & \textbf{100.0} & \textbf{99.9} & \textbf{98.4} & \cellcolor{gray!8}\textbf{93.7} & \cellcolor{gray!8}\textbf{97.8} & \cellcolor{gray!8}\textbf{98.6} & \cellcolor{gray!8}\textbf{98.8} & \textbf{98.4} \\
\bottomrule
\end{tabular}%
}
\end{table*}

\begin{table*}[t]
\centering
\caption{\textbf{\ourmethod{} outperforms all baselines under every defense, with gains of +20--34 points on average.} Attack success rate (ASR\%, higher is better) under different defenses, grouped by surrogate model. For NRP we report the average ASR\% over evaluated target models.}
\label{tab:asr_under_defenses}
\scriptsize
\setlength{\tabcolsep}{4pt}
\renewcommand{\arraystretch}{0.92}

\begin{tabular*}{\textwidth}{@{\extracolsep{\fill}} llccccc}
\toprule
\textbf{Surrogate} & \textbf{Attack Method} & \multicolumn{5}{c}{\textbf{Defense Methods}} \\
\cmidrule(lr){3-7}
& & \textbf{AT} & \textbf{HGD} & \textbf{NRP} & \textbf{RS} & \textbf{DiffPure} \\
\midrule

\multirow{8}{*}{ResNet-50}
& P2FA        & 44.8 & 67.5 & 52.8 & 32.9 & 21.9 \\
& BFA         & 44.8 & 77.4 & 57.7 & 32.8 & 26.9 \\
& MFAA        & 42.8 & 49.0 & 46.3 & 30.8 & 18.9 \\
& ANDA        & 42.7 & 66.1 & 45.7 & 31.4 & 28.0 \\
& GI-FGSM     & 40.9 & 25.2 & 37.7 & 29.1 & 15.9 \\
& ILPD        & 43.4 & 57.3 & 49.6 & 32.9 & 29.4 \\
& DiffAttack  & 46.5 & 38.5 & 47.0 & 40.6 & 35.5 \\
\cmidrule(lr){2-7}
& \ourmethod{} (Ours) & \textbf{57.6} & \textbf{91.3} & \textbf{77.3} & \textbf{64.1} & \textbf{63.9} \\
\midrule

\multirow{8}{*}{ResNet-152}
& P2FA        & 46.6 & 87.6 & 53.6 & 37.0 & 32.0 \\
& BFA         & 45.0 & 87.9 & 60.0 & 35.1 & 39.3 \\
& MFAA        & 44.4 & 64.3 & 46.6 & 33.3 & 26.5 \\
& ANDA        & 43.2 & 79.1 & 49.5 & 32.1 & 36.6 \\
& GI-FGSM     & 41.9 & 42.9 & 40.3 & 29.3 & 22.6 \\
& ILPD        & 44.2 & 65.9 & 52.3 & 34.8 & 41.6 \\
& DiffAttack  & 47.9 & 51.3 & 53.4 & 42.5 & 44.8 \\
\cmidrule(lr){2-7}
& \ourmethod{} (Ours) & \textbf{58.3} & \textbf{95.0} & \textbf{81.0} & \textbf{62.4} & \textbf{73.1} \\
\midrule

\multirow{8}{*}{VGG-16}
& P2FA        & 41.3 & 44.8 & 38.7 & 29.0 & 16.5 \\
& BFA         & 44.2 & 72.9 & 45.3 & 31.1 & 19.6 \\
& MFAA        & 43.7 & 51.5 & 38.9 & 32.9 & 22.0 \\
& ANDA        & 43.0 & 67.3 & 36.9 & 29.9 & 20.0 \\
& GI-FGSM     & 43.3 & 37.9 & 35.8 & 29.1 & 17.1 \\
& ILPD        & 42.9 & 67.9 & 43.2 & 31.2 & 24.6 \\
& DiffAttack  & 54.0 & 61.2 & 59.8 & 49.5 & 41.4 \\
\cmidrule(lr){2-7}
& \ourmethod{} (Ours)  & \textbf{74.4} & \textbf{99.4} & \textbf{93.7} & \textbf{82.0} & \textbf{88.1} \\
\bottomrule
\end{tabular*}
\end{table*}

\subsection{Comparison of Transferability}

We compare the transferability of \ourmethod{} against prior transfer attacks across three surrogate models (RN50, RN152, VGG16). For each surrogate, we generate adversarial examples on the surrogate and evaluate ASR on a diverse set of unseen target architectures, including both CNNs and vision transformers.

\paragraph{\textbf{Attacking unseen targets (CNNs and ViTs).}}

Table~\ref{tab:asr_per_model} reports transfer ASR across the full target suite.
Across all three surrogates, \ourmethod{} achieves the strongest overall transfer, with average ASR of
$89.9$ (RN50), $92.4$ (RN152), and $98.4$ (VGG16).
Relative to the best baseline per surrogate, \ourmethod{} improves average ASR by
$+6.5$ points for RN50 (vs.\ P2FA at $83.4$), $+1.8$ points for RN152 (vs.\ P2FA at $90.6$), and
$+19.1$ points for VGG16 (vs.\ ILPD at $79.3$).

The gains are most pronounced on transformer targets (ViT-B/16, PiT-B, Visformer-S, Swin-T): averaged over these four models,
\ourmethod{} improves ASR by $+13.7$ points when using RN50 as surrogate ($86.5$ vs.\ $72.8$ for the best baseline) and by
$+9.2$ points with RN152 ($91.7$ vs.\ $82.5$).
For example, with RN50 as surrogate, \ourmethod{} reaches $71.3$ ASR on ViT-B/16, exceeding the best baseline (BFA) by $+21.8$ points.
We attribute this to the fact that many baselines are designed around CNN feature hierarchies or pixel-space gradient signals, which overfit architecture-specific inductive biases.
In contrast, optimizing in VAE latent space biases perturbations toward low-frequency, spatially coherent changes that better align with features shared across CNNs and ViTs.

The particularly strong VGG16 surrogate results ($98.4$ average) likely reflect VGG's simpler, more generic feature representations: latent-space perturbations that exploit these broadly shared features transfer almost universally. Diffusion-based attacks such as DiffAttack, while also leveraging a generative prior, exhibit limited transfer in our setting, suggesting that strong transfer requires not only a generative prior but also an optimization procedure that produces perturbations robust to cross-architecture representation shifts.


\paragraph{\textbf{Attacking defense pipelines.}}

We next evaluate transferability under defense mechanisms (AT, HGD, NRP, RS, DiffPure).
As shown in Table~\ref{tab:asr_under_defenses}, \ourmethod{} consistently attains the highest ASR across all defenses and surrogates.
Averaged in the five defenses, \ourmethod{} achieves $70.8$ (RN50), $74.0$ (RN152) and $87.5$ (VGG16),
while the strongest baseline averages $47.9$, $53.5$, and $53.2$, respectively, improvements of $+22.9$, $+20.5$, and $+34.3$ points.
The gains are especially large against purification-based defenses (HGD, NRP, DiffPure), which attempt to remove adversarial signals via denoising or diffusion. Because \ourmethod{}'s perturbations are predominantly low-frequency and structurally aligned with the image content, they are harder to separate from the clean signal, making purification less effective.


\begin{figure}[t]
\centering
\includegraphics[scale=0.55]{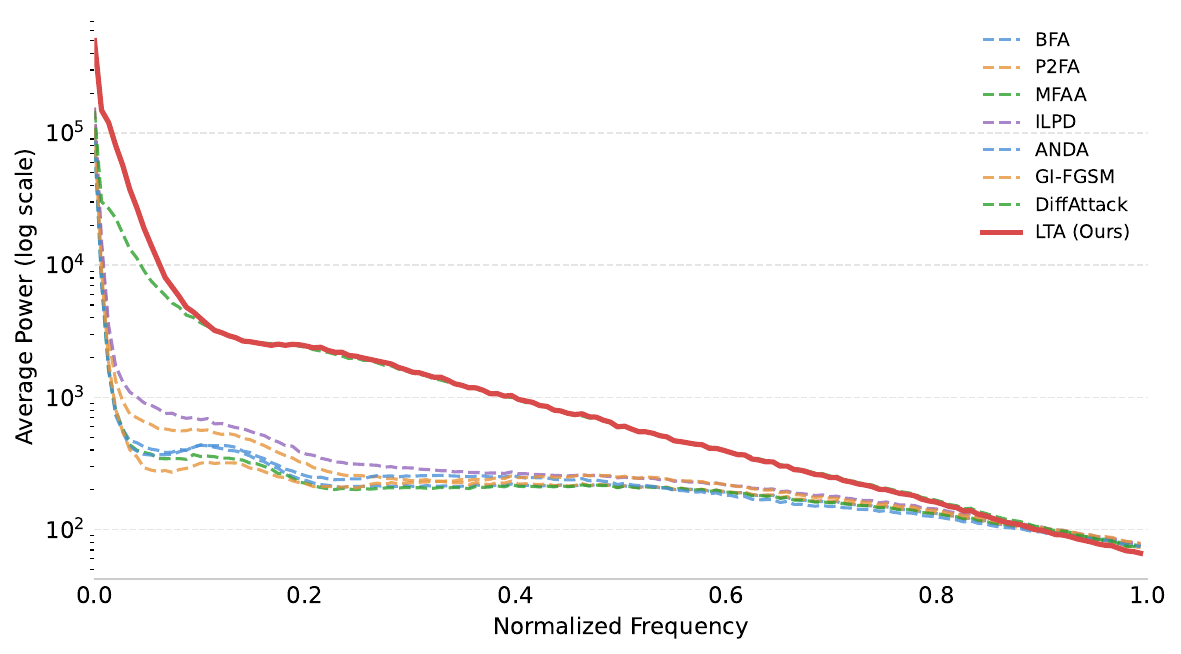}
\caption{\textbf{\ourmethod{} concentrates perturbation energy in low frequencies far more than any baseline.} Radially-averaged power spectrum of adversarial perturbations (log scale), averaged over 100 images. \ourmethod{} shows a steeper spectral roll-off, indicating stronger low-frequency bias.}
\label{fig:radial_spectrum}
\end{figure}

\subsection{User Study}

To complement automatic quality metrics, we conduct a user study to assess whether adversarial images appear visually modified.
We sample 50 adversarial images from each method (P2FA, GI-FGSM, DiffAttack, and \ourmethod{}) and 50 clean images from the same dataset.
Eight participants are each shown a single image at a time (randomized order) and answer a binary question: \emph{``Is this image original or modified?''}.
We report the \emph{fooling rate}, i.e., the fraction of trials in which an image is judged as \emph{original} (chance level is 50\%).
Clean images achieve a high fooling rate (91.5\%), confirming that participants understand the task.

\begin{figure}[t]
\centering
\includegraphics[scale=0.55]{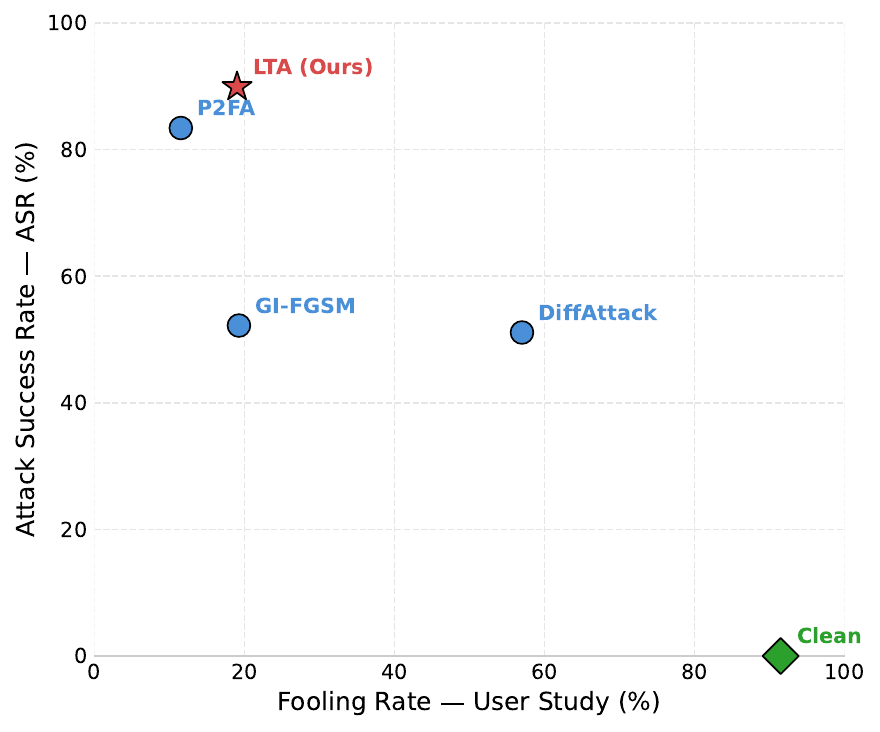}
\caption{\textbf{Attack strength vs.\ perceptual detectability.} Scatter plot of transfer ASR (\%, higher is better) versus user-study fooling rate (\%, higher means less detectable), comparing clean images and adversarial images from different methods.}
\label{fig:asr_vs_fooling}
\end{figure}

Among attacks, DiffAttack is the least detectable (57.0\%), while pixel-space baselines are detected more reliably (P2FA: 11.5\%, GI-FGSM: 19.2\%).
\ourmethod{} achieves a fooling rate of 19.0\%, comparable to strong pixel-space baselines.
Fig.~\ref{fig:asr_vs_fooling} plots transfer ASR against fooling rate, directly illustrating the strength, quality trade-off:
DiffAttack occupies a high-fooling/low-ASR regime, whereas \ourmethod{} attains substantially higher ASR at a similar detectability level to pixel-space attacks.
We note that this study is small-scale (8 participants); we report it as a directional signal rather than a definitive perceptual evaluation, and provide standard image-quality metrics in Supp~\ref{sec:image_quality_metrics} for a more comprehensive assessment.

\subsection{Perturbation Frequency Analysis}
\label{sec:freq_analysis}

To understand why \ourmethod{} transfers strongly across architectures, we analyze the spectral structure of the induced perturbations.
Given a clean image $x$ and its adversarial counterpart $x_{\text{adv}}$, we define the perturbation as $\delta = x_{\text{adv}} - x$.
All analyses in this subsection are computed on a random subset of 100 images drawn from our 1,000-image dataset.
For each method, we compute the 2D Fourier transform per-channel and analyze the (i) radially-averaged power spectrum, (ii) energy mass across coarse frequency bands,
and (iii) qualitative spatial patterns of the perturbation.


\paragraph{\textbf{Radial power spectrum.}}
Fig.~\ref{fig:radial_spectrum} reports the radially-averaged power spectrum of $\delta$ (log scale) as a function of normalized frequency ($0$=DC, $1$=Nyquist). For each image we compute $F(\delta)$ via a 2D FFT (with DC centered), form the power $|F(\delta)|^2$, and average over annuli of equal radial distance.
Across all baselines, \ourmethod{} exhibits a markedly \emph{more low-frequency dominated} profile,
characterized by a stronger concentration near DC and a steeper roll-off as frequency increases.  While methods differ in total perturbation magnitude, the spectral \emph{shape} reveals how energy is distributed: \ourmethod{} allocates relatively less energy to mid/high frequencies compared to pixel-space gradient baselines, which retain broader high-frequency content.

\begin{figure}[t]
\centering
\includegraphics[scale=0.25]{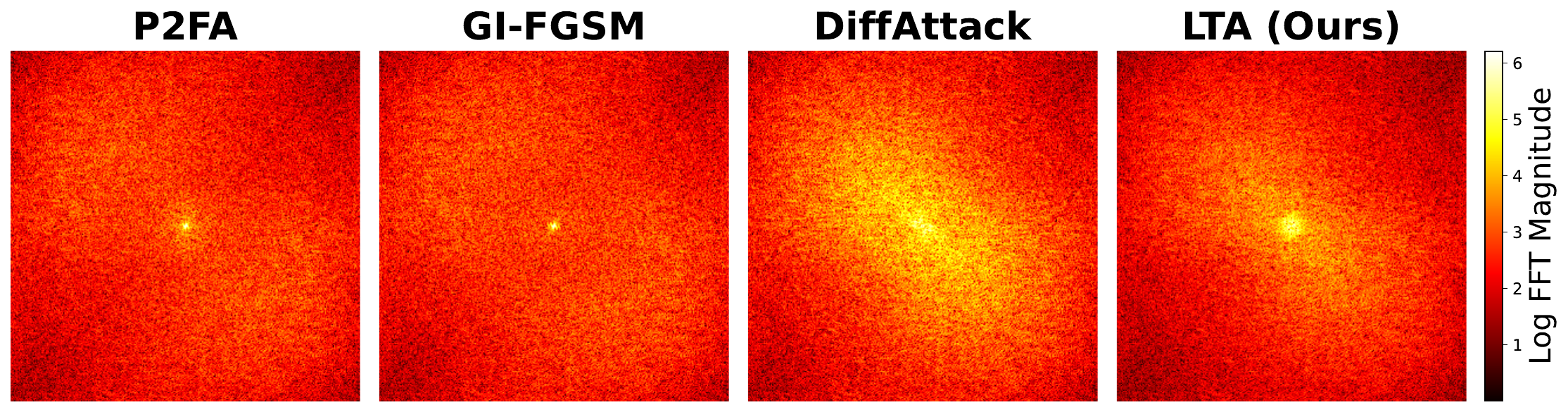}
\caption{\textbf{\ourmethod{}'s perturbation spectrum is tightly concentrated near DC, confirming its low-frequency bias.} Log-magnitude of the 2D FFT of $\delta=x_{\text{adv}}-x$ (DC centered), averaged over 100 images. Pixel-space baselines show energy spread to higher frequencies.}
\label{fig:fft_2d}
\end{figure}
\begin{figure}[t]
\centering
\includegraphics[width=\linewidth]{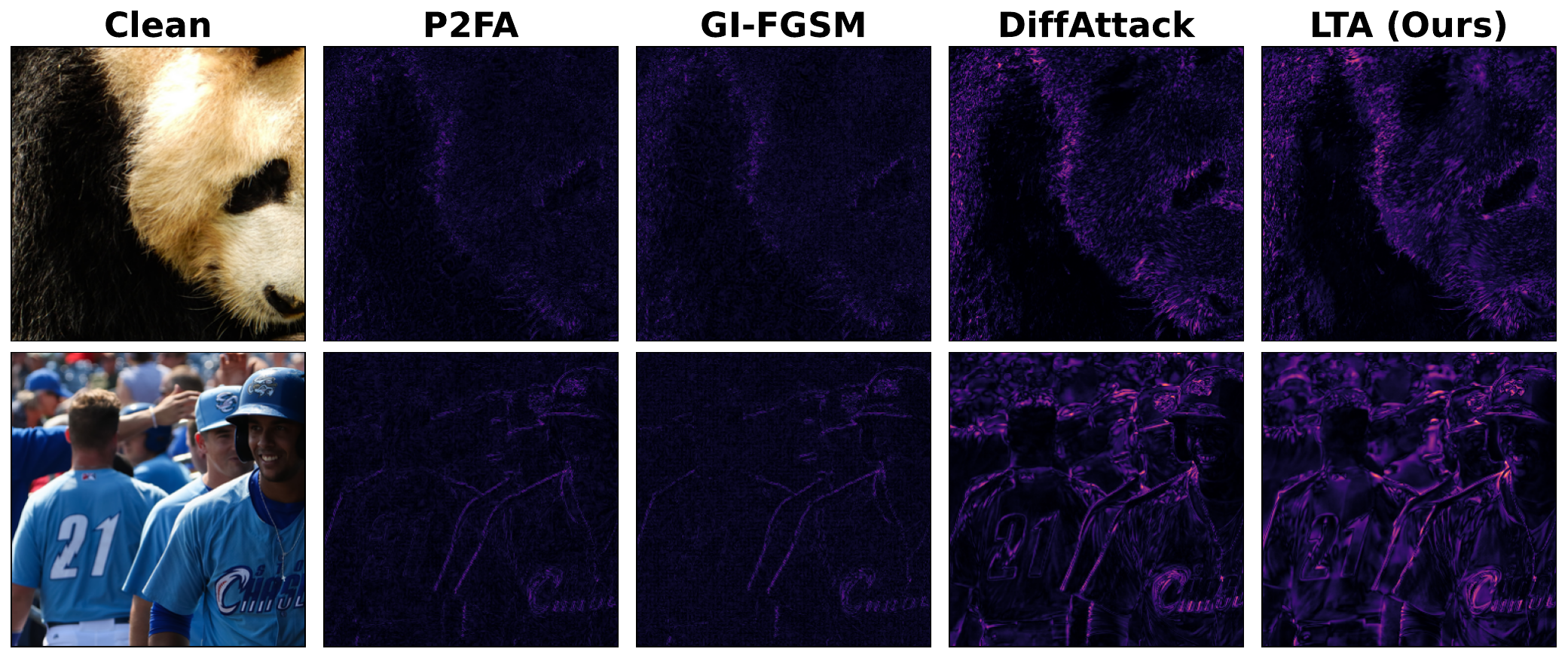}
\caption{\textbf{\ourmethod{} produces spatially coherent, structure-aligned perturbations rather than diffuse texture-like noise.} Per-pixel perturbation magnitude $\|\delta(p)\|_2$ for a representative image (clean image shown for reference; values normalized).}
\label{fig:perturb_vis}
\end{figure}

\paragraph{\textbf{2D FFT and spatial visualization.}}
Fig.~\ref{fig:fft_2d} shows the log-magnitude of the 2D Fourier spectrum of $\delta$ (DC centered) for each method, corroborating the radial analysis: \ourmethod{}'s spectrum is concentrated near the origin, while pixel-space baselines (P2FA, GI-FGSM) show energy distributed at larger radii.
Fig.~\ref{fig:perturb_vis} visualizes the per-pixel perturbation magnitude
\begin{equation*}
m(p) \;=\; \|\delta(p)\|_2 \;=\; \sqrt{\delta_r(p)^2 + \delta_g(p)^2 + \delta_b(p)^2}, \qquad \delta = x_{\text{adv}}-x,
\end{equation*}
with values normalized for visualization.
Compared to pixel-space baselines, which exhibit diffuse, texture-like patterns, \ourmethod{} produces a spatially coherent,
structure-aligned perturbation that concentrates on semantically salient regions (e.g., the object) rather than being uniformly spread over the background.
These spatial and spectral observations are consistent: the VAE decoder channels adversarial energy into structured, low-frequency directions.
Additional visualizations for all methods are provided in Supp~\ref{sec:supp_extra_vis}.

\subsection{Ablation Study}

We ablate the main components of \ourmethod{} using a fixed surrogate (ResNet-50) under the protocol in Sec.~\ref{sec:exp_setup}.
Table~\ref{tab:ablation_combined} reports both (i) \emph{transferability}, measured by average attack success rate (ASR\%) over unseen target architectures,
and (ii) \emph{perceptual quality}, measured by PSNR/SSIM (higher is better) and LPIPS/FID (lower is better).

The results reveal a clear separation between components that primarily improve transfer and those that primarily improve visual fidelity.
First, passing images through the VAE without any optimization (\textsc{VAE}) yields negligible attack success (Avg.\ ASR $10.2$), confirming that adversarial signal
must be explicitly induced by optimizing the latent code. Latent-space optimization alone (\textsc{Z-Opt}) already provides strong transfer (Avg.\ ASR $96.8$),
indicating that the latent parameterization is sufficiently expressive for transferable attacks.

Second, EOT is the main driver of transfer: adding EOT to latent optimization (\textsc{Z-Opt}+E) improves the average ASR from $96.8$ to $98.1$,
and combining EOT with the soft pixel-space constraint (\textsc{Z-Opt}+E+$\epsilon$) attains the best transfer ($98.2$) with a comparable quality profile.
This supports the role of EOT in mitigating resolution and preprocessing mismatch between the VAE decoder output and downstream classifiers.

Third, the soft pixel-space $\ell_\infty$ penalty acts primarily as a quality regularizer: \textsc{Z-Opt}+$\epsilon$ improves PSNR/SSIM
($19.87\!\rightarrow\!20.24$, $0.614\!\rightarrow\!0.624$) with minimal effect on ASR ($96.8$ vs.\ $96.5$), suggesting that softly penalizing post-decoding
budget violations improves fidelity without substantially restricting adversarial directions.

Finally, latent smoothing exposes the quality--transfer trade-off most directly. Smoothing improves perceptual metrics substantially (\textsc{Z-Opt}+S: PSNR $22.98$,
SSIM $0.767$, FID $85.00$) but sharply reduces transfer (Avg.\ ASR $82.0$). Adding smoothing on top of transfer-enhancing components recovers part of the loss
(\textsc{Z-Opt}+E+S: ASR $90.3$) while preserving improved quality.
We adopt \ourmethod{} (E+$\epsilon$+S) as our default configuration: it provides a balanced operating point on the transfer--quality trade-off, whereas EOT-only variants maximize ASR at a substantial perceptual cost.

\begin{figure}[t]
\centering
\includegraphics[width=\linewidth]{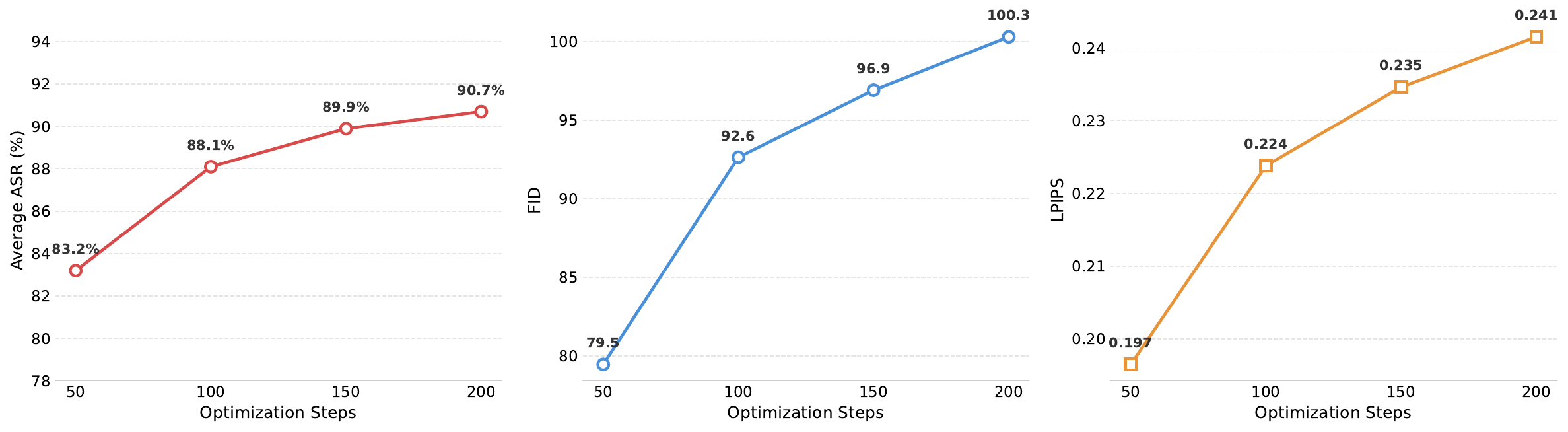}
\caption{\textbf{More optimization steps improve transfer at the cost of perceptual quality.} Average transfer ASR (left), FID (middle), and LPIPS (right) as a function of latent optimization steps (ResNet-50 surrogate, 50$\to$200 steps).}
\label{fig:opt_steps}
\end{figure}

\paragraph{\textbf{Effect of optimization steps.}}
Fig.~\ref{fig:opt_steps} (ResNet-50 surrogate) illustrates the same trade-off along the optimization trajectory: increasing the number of latent optimization steps monotonically improves average transfer ASR but gradually degrades perceptual quality (higher LPIPS and FID), confirming that attack strength and image fidelity are competing objectives in this framework.

\begin{table*}[t]
\centering
\caption{\textbf{EOT drives transferability while smoothing and the soft $\ell_\infty$ penalty improve perceptual quality, revealing a clear transfer--quality trade-off.} Ablation study (ResNet-50 surrogate) varying three components: EOT~(E), soft pixel-space $\ell_\infty$ penalty~($\epsilon$), and periodic latent Gaussian smoothing~(S).}
\label{tab:ablation_combined}
\scriptsize
\setlength{\tabcolsep}{5pt}
\renewcommand{\arraystretch}{1.05}

\begin{tabular}{lccc ccccc}
\toprule
\textbf{Config} & \textbf{E} & $\boldsymbol{\epsilon}$ & \textbf{S} &
\textbf{ASR Avg.} $\uparrow$ & \textbf{PSNR} $\uparrow$ & \textbf{SSIM} $\uparrow$ & \textbf{LPIPS} $\downarrow$ & \textbf{FID} $\downarrow$ \\
\midrule
\textsc{VAE} (Enc$\rightarrow$Dec)       & -- & -- & -- & 10.2 & \textbf{25.98} & \textbf{0.8425} & \textbf{0.1183} & \textbf{34.57} \\
\textsc{Z-Opt}                          & -- & -- & -- & 96.8 & 19.87 & 0.6141 & 0.3694 & 122.08 \\
\textsc{Z-Opt}+$\epsilon$               & -- & \checkmark & -- & 96.5 & 20.24 & 0.6239 & 0.3633 & 119.88 \\
\textsc{Z-Opt}+E                        & \checkmark & -- & -- & 98.1 & 19.31 & 0.5889 & 0.3870 & 129.89 \\
\textsc{Z-Opt}+S                        & -- & -- & \checkmark & 82.0 & 22.98 & 0.7670 & 0.2042 & 85.00 \\
\textsc{Z-Opt}+E+S                      & \checkmark & -- & \checkmark & 90.3 & 22.06 & 0.7386 & 0.2365 & 97.73 \\
\textsc{Z-Opt}+E+$\epsilon$             & \checkmark & \checkmark & -- & \textbf{98.2} & 19.76 & 0.6014 & 0.3805 & 127.95 \\
\textsc{Z-Opt}+$\epsilon$+S             & -- & \checkmark & \checkmark & 81.9 & 23.09 & 0.7686 & 0.2027 & 83.78 \\
\midrule
\textbf{\ourmethod{}} (E+$\epsilon$+S)  & \checkmark & \checkmark & \checkmark & 89.9 & 22.26 & 0.7415 & 0.2346 & 96.91 \\
\bottomrule
\end{tabular}
\end{table*}

\section{Limitations}
\label{sec:limitations}

While \ourmethod{} demonstrates strong transferability of adversarial perturbations, it has several important limitations.

\paragraph{\textbf{Dependence on the VAE Prior.}}
Our approach relies on the latent space of a pretrained Stable Diffusion VAE as an implicit image prior.
Consequently, the space of attainable perturbations is restricted to those representable by the decoder.
Although this constraint promotes spatial coherence and suppresses high-frequency noise, it may also exclude adversarial directions that are effective but lie outside the VAE manifold.
This can reduce attack strength in scenarios where optimal perturbations require fine-grained, high-frequency pixel modifications.
\paragraph{\textbf{Computational Overhead.}}
Compared to conventional pixel-space attacks, \ourmethod{} incurs additional computational cost due to repeated VAE decoding, expectation over transformations, and periodic latent smoothing.
The need to sample multiple transformations per iteration further increases runtime and memory usage, limiting scalability to high-resolution images or large batch sizes.
For completeness, we provide a systematic wall-clock analysis (mean$\pm$std per image, same hardware) comparing all attacks in the supplementary material (Sec.~\ref{sec:attacks_runtime}).

\section{Conclusion}
\label{sec:conclusion}

We introduced \ourmethod{}, a transfer-based adversarial attack that optimizes perturbations in the latent space of a pretrained Stable Diffusion VAE rather than directly in pixel space. By leveraging the VAE decoder as an implicit image prior, our method generates adversarial examples that are spatially coherent and dominated by low-frequency structure, reducing the high-frequency artifacts commonly produced by pixel-space gradient attacks. Our approach integrates expectation over transformations and a soft pixel-space constraint to improve robustness to preprocessing and enhance cross-model transferability, while remaining compatible with standard optimization procedures. Empirical results demonstrate that latent-space adversarial optimization can achieve superior attack success rates while producing perturbations that better align with the underlying image manifold. This work highlights the potential of generative model latent spaces as a principled alternative domain for adversarial optimization. Beyond improving transferability, our framework provides a step toward unifying adversarial attacks with learned image priors, suggesting new directions for studying robustness under structured and perceptually grounded perturbations. Future work may explore richer generative priors, stricter constraint enforcement, and extensions to multimodal or video-based adversarial settings.

\clearpage  


%
%
\bibliographystyle{splncs04}
\bibliography{references}

@String(CVPR  = {IEEE Conf. Comput. Vis. Pattern Recog.})

@String(ICCV  = {Int. Conf. Comput. Vis.})

@String(ECCV  = {Eur. Conf. Comput. Vis.})

@String(NeurIPS = {Adv. Neural Inform. Process. Syst.})

@String(ICML  = {Int. Conf. Mach. Learn.})

@String(ICLR  = {Int. Conf. Learn. Represent.})

@String(CVPRW = {IEEE Conf. Comput. Vis. Pattern Recog. Worksh.})

@String(IJCAI = {IJCAI})

@String(CVPR  = {CVPR})

@String(ICCV  = {ICCV})

@String(ECCV  = {ECCV})

@String(NeurIPS = {NeurIPS})

@String(ICML  = {ICML})

@String(ICLR  = {ICLR})

@String(CVPRW = {CVPRW})

@inproceedings{liu2025pixel2feature,
  title={Pixel2Feature attack (P2FA): Rethinking the perturbed space to enhance adversarial transferability},
  author={Liu, Renpu and Wu, Hao and Zhang, Jiawei and Cheng, Xin and Luo, Xiangyang and Ma, Bin and Wang, Jinwei},
  booktitle={Forty-second International Conference on Machine Learning},
  year={2025}
}

@article{wang2024improving,
  title={Improving the transferability of adversarial examples through black-box feature attacks},
  author={Wang, Maoyuan and Wang, Jinwei and Ma, Bin and Luo, Xiangyang},
  journal={Neurocomputing},
  volume={595},
  pages={127863},
  year={2024},
  publisher={Elsevier}
}

@article{zheng2025enhancing,
  title={Enhancing the transferability of adversarial attacks via multi-feature attention},
  author={Zheng, Desheng and Ke, Wuping and Li, Xiaoyu and Duan, Yaoxin and Yin, Guangqiang and Min, Fan},
  journal={IEEE Transactions on Information Forensics and Security},
  volume={20},
  pages={1462--1474},
  year={2025},
  publisher={IEEE}
}

@inproceedings{fang2024strong,
  title={Strong transferable adversarial attacks via ensembled asymptotically normal distribution learning},
  author={Fang, Zhengwei and Wang, Rui and Huang, Tao and Jing, Liping},
  booktitle={Proceedings of the IEEE/CVF Conference on Computer Vision and Pattern Recognition},
  pages={24841--24850},
  year={2024}
}

@misc{wang2024boostingtransferabilityadversarialattacks,
      title={Boosting the Transferability of Adversarial Attacks with Global Momentum Initialization}, 
      author={Jiafeng Wang and Zhaoyu Chen and Kaixun Jiang and Dingkang Yang and Lingyi Hong and Pinxue Guo and Haijing Guo and Wenqiang Zhang},
      year={2024},
      eprint={2211.11236},
      archivePrefix={arXiv},
      primaryClass={cs.CV},
      url={https://arxiv.org/abs/2211.11236}, 
}

@article{li2023improving,
  title={Improving adversarial transferability via intermediate-level perturbation decay},
  author={Li, Qizhang and Guo, Yiwen and Zuo, Wangmeng and Chen, Hao},
  journal={Advances in Neural Information Processing Systems},
  volume={36},
  pages={32900--32912},
  year={2023}
}

@article{chen2024diffusion,
  title={Diffusion models for imperceptible and transferable adversarial attack},
  author={Chen, Jianqi and Chen, Hao and Chen, Keyan and Zhang, Yilan and Zou, Zhengxia and Shi, Zhenwei},
  journal={IEEE Transactions on Pattern Analysis and Machine Intelligence},
  volume={47},
  number={2},
  pages={961--977},
  year={2024},
  publisher={IEEE}
}

@inproceedings{he2016deep,
  title={Deep residual learning for image recognition},
  author={He, Kaiming and Zhang, Xiangyu and Ren, Shaoqing and Sun, Jian},
  booktitle={Proceedings of the IEEE conference on computer vision and pattern recognition},
  pages={770--778},
  year={2016}
}

@article{simonyan2014very,
  title={Very deep convolutional networks for large-scale image recognition},
  author={Simonyan, Karen and Zisserman, Andrew},
  journal={arXiv preprint arXiv:1409.1556},
  year={2014}
}

@inproceedings{sandler2018mobilenetv2,
  title={Mobilenetv2: Inverted residuals and linear bottlenecks},
  author={Sandler, Mark and Howard, Andrew and Zhu, Menglong and Zhmoginov, Andrey and Chen, Liang-Chieh},
  booktitle={Proceedings of the IEEE conference on computer vision and pattern recognition},
  pages={4510--4520},
  year={2018}
}

@inproceedings{szegedy2016rethinking,
  title={Rethinking the inception architecture for computer vision},
  author={Szegedy, Christian and Vanhoucke, Vincent and Ioffe, Sergey and Shlens, Jon and Wojna, Zbigniew},
  booktitle={Proceedings of the IEEE conference on computer vision and pattern recognition},
  pages={2818--2826},
  year={2016}
}

@article{dosovitskiy2020image,
  title={An image is worth 16x16 words: Transformers for image recognition at scale},
  author={Dosovitskiy, Alexey and Beyer, Lucas and Kolesnikov, Alexander and Weissenborn, Dirk and Zhai, Xiaohua and Unterthiner, Thomas and Dehghani, Mostafa and Minderer, Matthias and Heigold, Georg and Gelly, Sylvain and others},
  journal={arXiv preprint arXiv:2010.11929},
  year={2020}
}

@inproceedings{heo2021rethinking,
  title={Rethinking spatial dimensions of vision transformers},
  author={Heo, Byeongho and Yun, Sangdoo and Han, Dongyoon and Chun, Sanghyuk and Choe, Junsuk and Oh, Seong Joon},
  booktitle={Proceedings of the IEEE/CVF international conference on computer vision},
  pages={11936--11945},
  year={2021}
}

@inproceedings{chen2021visformer,
  title={Visformer: The vision-friendly transformer},
  author={Chen, Zhengsu and Xie, Lingxi and Niu, Jianwei and Liu, Xuefeng and Wei, Longhui and Tian, Qi},
  booktitle={Proceedings of the IEEE/CVF international conference on computer vision},
  pages={589--598},
  year={2021}
}

@inproceedings{liu2021swin,
  title={Swin transformer: Hierarchical vision transformer using shifted windows},
  author={Liu, Ze and Lin, Yutong and Cao, Yue and Hu, Han and Wei, Yixuan and Zhang, Zheng and Lin, Stephen and Guo, Baining},
  booktitle={Proceedings of the IEEE/CVF international conference on computer vision},
  pages={10012--10022},
  year={2021}
}

@article{tramer2017ensemble,
  title={Ensemble adversarial training: Attacks and defenses},
  author={Tram{\`e}r, Florian and Kurakin, Alexey and Papernot, Nicolas and Goodfellow, Ian and Boneh, Dan and McDaniel, Patrick},
  journal={arXiv preprint arXiv:1705.07204},
  year={2017}
}

@inproceedings{liao2018defense,
  title={Defense against adversarial attacks using high-level representation guided denoiser},
  author={Liao, Fangzhou and Liang, Ming and Dong, Yinpeng and Pang, Tianyu and Hu, Xiaolin and Zhu, Jun},
  booktitle={Proceedings of the IEEE conference on computer vision and pattern recognition},
  pages={1778--1787},
  year={2018}
}

@inproceedings{cohen2019certified,
  title={Certified adversarial robustness via randomized smoothing},
  author={Cohen, Jeremy and Rosenfeld, Elan and Kolter, Zico},
  booktitle={international conference on machine learning},
  pages={1310--1320},
  year={2019},
  organization={PMLR}
}

@inproceedings{naseer2020self,
  title={A self-supervised approach for adversarial robustness},
  author={Naseer, Muzammal and Khan, Salman and Hayat, Munawar and Khan, Fahad Shahbaz and Porikli, Fatih},
  booktitle={Proceedings of the IEEE/CVF conference on computer vision and pattern recognition},
  pages={262--271},
  year={2020}
}

@article{nie2022diffusion,
  title={Diffusion models for adversarial purification},
  author={Nie, Weili and Guo, Brandon and Huang, Yujia and Xiao, Chaowei and Vahdat, Arash and Anandkumar, Anima},
  journal={arXiv preprint arXiv:2205.07460},
  year={2022}
}

@inproceedings{szegedy2014intriguing,
  title={Intriguing properties of neural networks},
  author={Szegedy, Christian and Zaremba, Wojciech and Sutskever, Ilya and et al.},
  booktitle={ICLR},
  year={2014}
}

@inproceedings{goodfellow2015explaining,
  title={Explaining and harnessing adversarial examples},
  author={Goodfellow, Ian and Shlens, Jonathon and Szegedy, Christian},
  booktitle={ICLR},
  year={2015}
}

@inproceedings{kurakin2017adversarial,
  title={Adversarial examples in the physical world},
  author={Kurakin, Alexey and Goodfellow, Ian and Bengio, Samy},
  booktitle={ICLR Workshop},
  year={2017}
}

@inproceedings{carlini2017towards,
  title={Towards evaluating the robustness of neural networks},
  author={Carlini, Nicholas and Wagner, David},
  booktitle={IEEE Symposium on Security and Privacy},
  year={2017}
}

@inproceedings{kingma2014autoencoding,
  title={Auto-Encoding Variational Bayes},
  author={Kingma, Diederik P and Welling, Max},
  booktitle={ICLR},
  year={2014}
}

@inproceedings{rezende2014stochastic,
  title={Stochastic Backpropagation and Approximate Inference in Deep Generative Models},
  author={Rezende, Danilo J and Mohamed, Shakir and Wierstra, Daan},
  booktitle={ICML},
  year={2014}
}

@article{ilyas2019adversarial,
  title={Adversarial examples are not bugs, they are features},
  author={Ilyas, Andrew and Santurkar, Shibani and Tsipras, Dimitris and Engstrom, Logan and Tran, Brandon and Madry, Aleksander},
  journal={Advances in Neural Information Processing Systems (NeurIPS)},
  year={2019}
}

@article{tsipras2019robustness,
  title={Robustness may be at odds with accuracy},
  author={Tsipras, Dimitris and Santurkar, Shibani and Engstrom, Logan and Turner, Alexander and Madry, Aleksander},
  journal={International Conference on Learning Representations (ICLR)},
  year={2019}
}

@article{sharma2019highfrequency,
  title={On the importance of high-frequency components in adversarial images},
  author={Sharma, Yash and Chen, Pin-Yu and Yi, Jinfeng and others},
  journal={arXiv preprint arXiv:1910.04715},
  year={2019}
}

@inproceedings{papernot2016transferability,
  title={Transferability in machine learning: from phenomena to black-box attacks},
  author={Papernot, Nicolas and McDaniel, Patrick and Goodfellow, Ian},
  booktitle={USENIX Security Symposium},
  year={2017}
}

@inproceedings{dong2018boosting,
  title={Boosting adversarial attacks with momentum},
  author={Dong, Yinpeng and Liao, Fangzhou and Pang, Tianyu and Su, Hang and Hu, Xiaolin and Li, Jianguo and Zhu, Jun},
  booktitle={CVPR},
  year={2018}
}

@inproceedings{xie2019improving,
  title={Improving transferability of adversarial examples with input diversity},
  author={Xie, Cihang and Zhang, Zhishuai and Zhou, Yuyin and Bai, Song and Wang, Jianyu and Ren, Zhou and Yuille, Alan},
  booktitle={CVPR},
  year={2019}
}

@inproceedings{athalye2018obfuscated,
  title={Obfuscated gradients give a false sense of security},
  author={Athalye, Anish and Carlini, Nicholas and Wagner, David},
  booktitle={ICML},
  year={2018}
}

@inproceedings{song2018constructing,
  title={Constructing unrestricted adversarial examples with generative models},
  author={Song, Yang and Shu, Rui and Kushman, Nate and Ermon, Stefano},
  booktitle={NeurIPS},
  year={2018}
}

@inproceedings{schott2019towards,
  title={Towards the first adversarially robust neural network model on MNIST},
  author={Schott, Lukas and Rauber, Jonas and Bethge, Matthias and Brendel, Wieland},
  booktitle={ICLR},
  year={2019}
}

@article{joshi2020towards,
  title={Towards realistic adversarial examples using generative models},
  author={Joshi, Shashank and Liu, Jiaming and Gu, Jian and others},
  journal={CVPR},
  year={2020}
}

@inproceedings{naseer2021generating,
  title={Generating adversarial examples via latent space manipulation},
  author={Naseer, Muzammal and Ranasinghe, Kanchana and Khan, Salman and Hayat, Munawar and Khan, Fahad Shahbaz},
  booktitle={CVPR},
  year={2021}
}

@article{huang2023advdiffusion,
  title={Diffusion-based adversarial attacks},
  author={Huang, Huiling and others},
  journal={arXiv preprint arXiv:2303.01673},
  year={2023}
}

@inproceedings{guo2018countering,
  title={Countering adversarial images using input transformations},
  author={Guo, Chuan and Rana, Mayank and Cisse, Moustapha and van der Maaten, Laurens},
  booktitle={International Conference on Learning Representations (ICLR)},
  year={2018}
}

@inproceedings{rozsa2016adversarial,
  title={Adversarial diversity and hard positive generation},
  author={Rozsa, Andras and Rudd, Ethan M. and Boult, Terrance E.},
  booktitle={Proceedings of the IEEE Conference on Computer Vision and Pattern Recognition Workshops (CVPRW)},
  year={2016}
}

@inproceedings{engstrom2019rotation,
  title={A rotation and a translation suffice: Fooling CNNs with simple transformations},
  author={Engstrom, Logan and Tran, Brandon and Tsipras, Dimitris and Schmidt, Ludwig and Madry, Aleksander},
  booktitle={International Conference on Learning Representations (ICLR)},
  year={2019}
}

@inproceedings{gupta2023semantic,
  title={Semantic adversarial attacks},
  author={Gupta, Shivam and others},
  booktitle={IEEE/CVF Conference on Computer Vision and Pattern Recognition (CVPR)},
  year={2023}
}

@inproceedings{madry2018towards,
  title     = {Towards Deep Learning Models Resistant to Adversarial Attacks},
  author    = {Madry, Aleksander and Makelov, Aleksandar and Schmidt, Ludwig and Tsipras, Dimitris and Vladu, Adrian},
  booktitle = {International Conference on Learning Representations (ICLR)},
  year      = {2018}
}

@inproceedings{moosavi2016deepfool,
  title     = {DeepFool: A Simple and Accurate Method to Fool Deep Neural Networks},
  author    = {Moosavi-Dezfooli, Seyed-Mohsen and Fawzi, Alhussein and Frossard, Pascal},
  booktitle = {IEEE Conference on Computer Vision and Pattern Recognition (CVPR)},
  year      = {2016}
}

@article{tanay2016boundary,
  title   = {A Boundary Tilting Perspective on the Phenomenon of Adversarial Examples},
  author  = {Tanay, Thomas and Griffin, Lewis},
  journal = {arXiv preprint arXiv:1608.07690},
  year    = {2016}
}

@inproceedings{fawzi2018analysis,
  title     = {Analysis of Classifiers' Robustness to Adversarial Perturbations},
  author    = {Fawzi, Alhussein and Moosavi-Dezfooli, Seyed-Mohsen and Frossard, Pascal},
  booktitle = {International Conference on Machine Learning (ICML)},
  year      = {2018}
}

@inproceedings{sharma2019beyond,
  title     = {Beyond Pixel-Wise Attacks: Feature Space Adversaries},
  author    = {Sharma, Yash and Chen, Pin-Yu and Murdock, Caleb and Tygar, J. D.},
  booktitle = {Workshop on Security and Privacy in Machine Learning (NeurIPS)},
  year      = {2019}
}

@inproceedings{wu2020adversarial,
  title     = {Adversarial Examples in the Physical World},
  author    = {Wu, Tong and others},
  booktitle = {European Conference on Computer Vision (ECCV)},
  year      = {2020}
}

@inproceedings{dong2019evading,
  title     = {Evading Defenses to Transferable Adversarial Examples by Translation-Invariant Attacks},
  author    = {Dong, Yinpeng and Pang, Tianyu and Su, Hang and Zhu, Jun},
  booktitle = {IEEE Conference on Computer Vision and Pattern Recognition (CVPR)},
  year      = {2019}
}

@inproceedings{huang2021intermediate,
  title     = {Boosting Adversarial Transferability with Intermediate-Level Attack},
  author    = {Huang, Qian and Li, Jing and He, Ran},
  booktitle = {IEEE International Conference on Computer Vision (ICCV)},
  year      = {2021}
}

@inproceedings{liu2017delving,
  title     = {Delving into Transferable Adversarial Examples and Black-box Attacks},
  author    = {Liu, Yanpei and Chen, Xinyun and Liu, Chang and Song, Dawn},
  booktitle = {International Conference on Learning Representations (ICLR)},
  year      = {2017}
}

@inproceedings{advgan,
  title     = {AdvGAN: Generating Adversarial Examples via Generative Adversarial Networks},
  author    = {Xiao, Chaowei and Li, Bo and Zhu, Jun and He, Warren and Liu, Mingyan and Song, Dawn},
  booktitle = {International Joint Conference on Artificial Intelligence (IJCAI)},
  year      = {2018}
}

@inproceedings{poursaeed2018generative,
  title     = {Generative Adversarial Perturbations},
  author    = {Poursaeed, Omid and Katsman, Isay and Gao, Bicheng and Belongie, Serge},
  booktitle = {IEEE Conference on Computer Vision and Pattern Recognition (CVPR)},
  year      = {2018}
}

@article{papernot2017transferability,
  title   = {Transferability in Machine Learning: From Phenomena to Black-Box Attacks using Adversarial Samples},
  author  = {Papernot, Nicolas and McDaniel, Patrick and Goodfellow, Ian},
  journal = {arXiv preprint arXiv:1605.07277},
  year    = {2017}
}

@inproceedings{brendel2018decisionbased,
  title     = {Decision-Based Adversarial Attacks: Reliable Attacks Against Black-Box Machine Learning Models},
  author    = {Brendel, Wieland and Rauber, Jonas and Bethge, Matthias},
  booktitle = {International Conference on Learning Representations (ICLR)},
  year      = {2018}
}
\newpage

\clearpage
\vfill
\begin{center}
{\Large \textbf{Supplementary Material}}
\end{center}

\section{Image Quality Metrics}
\label{sec:image_quality_metrics}

Table~\ref{tab:image_quality_metrics_no_l2} reports perceptual metrics for adversarial examples generated with different surrogates.
Overall, \ourmethod{} produces perturbations with a distinctly different quality profile than pixel-space transfer baselines: its decoded perturbations are spatially smooth and
typically yield lower LPIPS than most gradient-based baselines, consistent with the low-frequency bias induced by latent-space optimization.
At the same time, methods that explicitly optimize for perceptual similarity (notably DiffAttack) achieve substantially better LPIPS and FID, reflecting a different operating point that prioritizes perceptual fidelity.
Across surrogates, \ourmethod{} attains competitive FID relative to non-diffusion baselines (e.g., comparable to ANDA/GI-FGSM for RN50/RN152), while maintaining strong transfer ASR (Table~\ref{tab:asr_per_model}),
highlighting a favorable effectiveness--quality trade-off for transfer-based attacks.

\begin{table*}[h]
\centering
\caption{Image quality metrics of adversarial examples, grouped by surrogate model.}
\label{tab:image_quality_metrics_no_l2}
\scriptsize
\renewcommand{\arraystretch}{0.95} 

\begin{tabular*}{\textwidth}{@{\extracolsep{\fill}} llcccc}
\toprule
\textbf{Surrogate} & \textbf{Method} & \textbf{PSNR} $\uparrow$ & \textbf{SSIM} $\uparrow$ & \textbf{LPIPS} $\downarrow$ & \textbf{FID} $\downarrow$ \\
\midrule

\multirow{8}{*}{ResNet-50}
 & P2FA       & 26.92 & \textbf{0.7648} & 0.2315 & 121.09 \\
 & BFA        & 26.88 & 0.7585 & 0.2710 & 111.18 \\
 & MFAA       & 27.00 & 0.7609 & 0.2700 & 106.03 \\
 & ANDA       & \textbf{27.02} & 0.7586 & 0.2848 & 99.99 \\
 & GI-FGSM     & 25.86 & 0.7180 & 0.3227 & 95.84 \\
 & ILPD       & 25.01 & 0.6945 & 0.3024 & 114.74 \\
 & DiffAttack & 23.31 & 0.7568 & \textbf{0.1240} & \textbf{48.96} \\
  \cmidrule(lr){2-6}
 & \ourmethod{} (Ours) & 22.26 & 0.7415 & 0.2346 & 96.91 \\
\midrule

\multirow{8}{*}{ResNet-152}
 & P2FA       & 26.61 & 0.7575 & 0.2562 & 121.92 \\
 & BFA        & 26.82 & 0.7578 & 0.2800 & 103.68 \\
 & MFAA       & 26.89 & 0.7587 & 0.2850 & 92.89 \\
 & ANDA       & \textbf{27.49} & \textbf{0.7763} & 0.2680 & 100.69 \\
 & GI-FGSM     & 25.83 & 0.7178 & 0.3361 & 99.96 \\
 & ILPD       & 24.94 & 0.6930 & 0.3174 & 115.54 \\
 & DiffAttack & 23.09 & 0.7510 & \textbf{0.1307} & \textbf{51.72} \\
  \cmidrule(lr){2-6}
 & \ourmethod{} (Ours) & 21.99 & 0.7351 & 0.2380 & 100.32 \\
\midrule

\multirow{8}{*}{VGG-16}
 & P2FA       & 26.12 & 0.7418 & 0.2622 & 138.83 \\
 & BFA        & 26.64 & 0.7569 & 0.2858 & 136.55 \\
 & MFAA       & 25.58 & 0.7183 & 0.3211 & 130.76 \\
 & ANDA       & \textbf{27.82} & \textbf{0.7891} & 0.2493 & 96.12 \\
 & GI-FGSM     & 26.01 & 0.7260 & 0.3177 & 100.57 \\
 & ILPD       & 24.89 & 0.6949 & 0.3045 & 138.27 \\
 & DiffAttack & 22.64 & 0.7397 & \textbf{0.1450} & \textbf{57.65} \\
 \cmidrule(lr){2-6}
 & \ourmethod{} (Ours) & 18.75 & 0.6183 & 0.3200 & 98.95 \\
\bottomrule
\end{tabular*}

\end{table*}

\section{\ourmethod{} Hyperparameter Analysis}
\label{sec:hparam_analysis}

We analyze the sensitivity of \ourmethod{} to key hyperparameters under the standard protocol (Sec.~\ref{sec:exp_setup}) using \textbf{ResNet-50} as the surrogate.
For each sweep we report the \emph{average transfer ASR} across target models together with two representative quality indicators, \emph{FID} and \emph{LPIPS}
(lower is better). Overall, the results expose a predictable effectiveness quality trade-off: stronger smoothing improves perceptual metrics but may reduce transfer.

\begin{figure}[h]
\centering
\includegraphics[width=\linewidth]{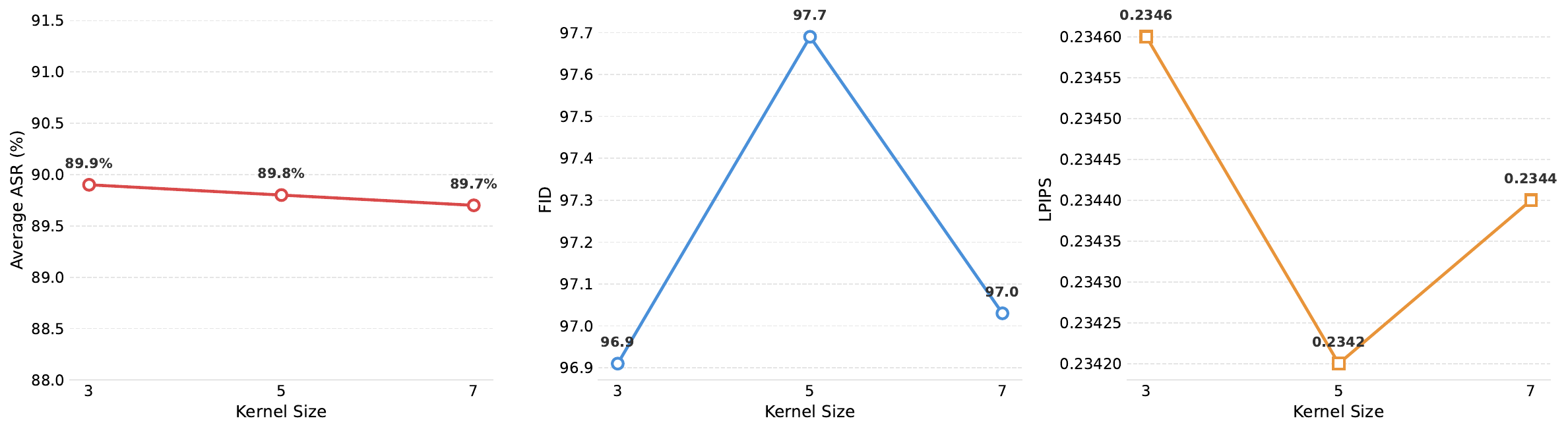}
\caption{\textbf{Effect of smoothing kernel size (ResNet-50 surrogate).}
Average transfer ASR (left) and quality metrics FID/LPIPS (middle/right) as a function of the Gaussian kernel size. Results are largely insensitive in the tested range.}
\label{fig:ablation_kernel_size}
\end{figure}

\paragraph{\textbf{Smoothing kernel size.}}
Fig.~\ref{fig:ablation_kernel_size} varies the Gaussian kernel size while keeping other settings fixed.
Performance is largely insensitive in the tested range: average ASR remains essentially unchanged ($\approx 89.7$ - $89.9$), with only minor differences in FID/LPIPS.
This suggests that the presence of smoothing matters more than the exact kernel support, and motivates using a small kernel for efficiency.

\begin{figure}[h]
\centering
\includegraphics[width=\linewidth]{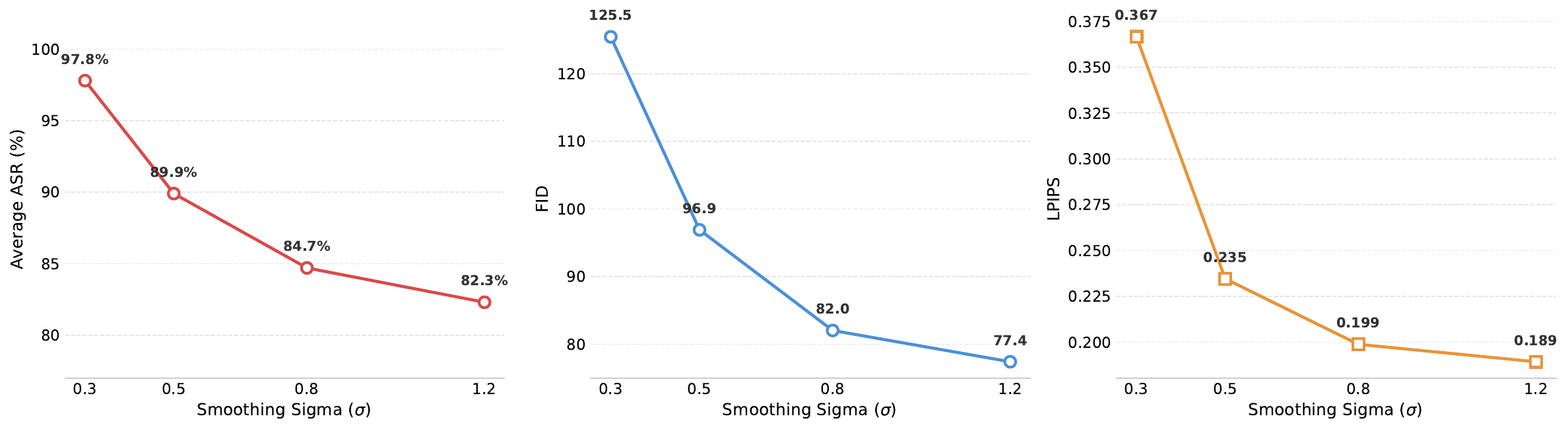}
\caption{\textbf{Effect of smoothing strength (ResNet-50 surrogate).}
Increasing Gaussian $\sigma$ improves perceptual quality (lower FID/LPIPS) but reduces transfer ASR, revealing a clear effectiveness--quality trade-off.}
\label{fig:ablation_sigma}
\end{figure}

\paragraph{\textbf{Smoothing strength ($\sigma$).}}
Fig.~\ref{fig:ablation_sigma} varies the Gaussian standard deviation.
As $\sigma$ increases, image quality improves substantially (lower FID and LPIPS), but transferability drops sharply
(e.g., average ASR decreases from $97.8$ at $\sigma{=}0.3$ to $82.3$ at $\sigma{=}1.2$).
This confirms that aggressive smoothing suppresses high-frequency artifacts but also removes attack-critical components.
In our experiments we adopt an intermediate setting ($\sigma{=}0.5$) as a balanced operating point.

\begin{figure}[h!]
\centering
\includegraphics[width=\linewidth]{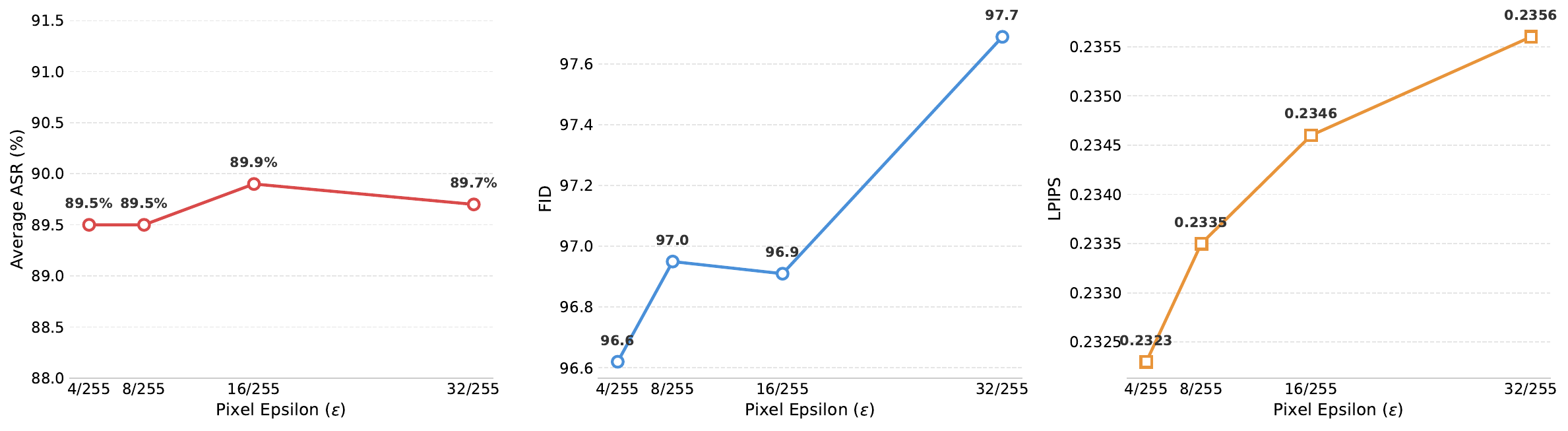}
\caption{\textbf{Effect of pixel budget (ResNet-50 surrogate).}
Average transfer ASR and quality metrics across different $\ell_\infty$ budgets. Transfer is relatively stable over the tested range, while quality slightly degrades for larger $\epsilon$.}
\label{fig:ablation_epsilon}
\end{figure}

\paragraph{\textbf{Pixel budget ($\epsilon$).}}
Fig.~\ref{fig:ablation_epsilon} sweeps the pixel-space $\ell_\infty$ budget. 
Across $\epsilon \in \{4,8,16,32\}/255$, average ASR remains stable ($\approx 89.5$ - $89.9$), while perceptual quality slightly degrades as the budget increases
(e.g., LPIPS and FID trend upward for larger $\epsilon$).
We therefore report results with $\epsilon{=}16/255$ as a standard setting, while noting that smaller budgets can yield comparable transfer with marginally improved perceptual metrics.

\section{User Study}
\label{sec:user_study}

\begin{figure}[t]
\centering
\includegraphics[width=0.7\linewidth]{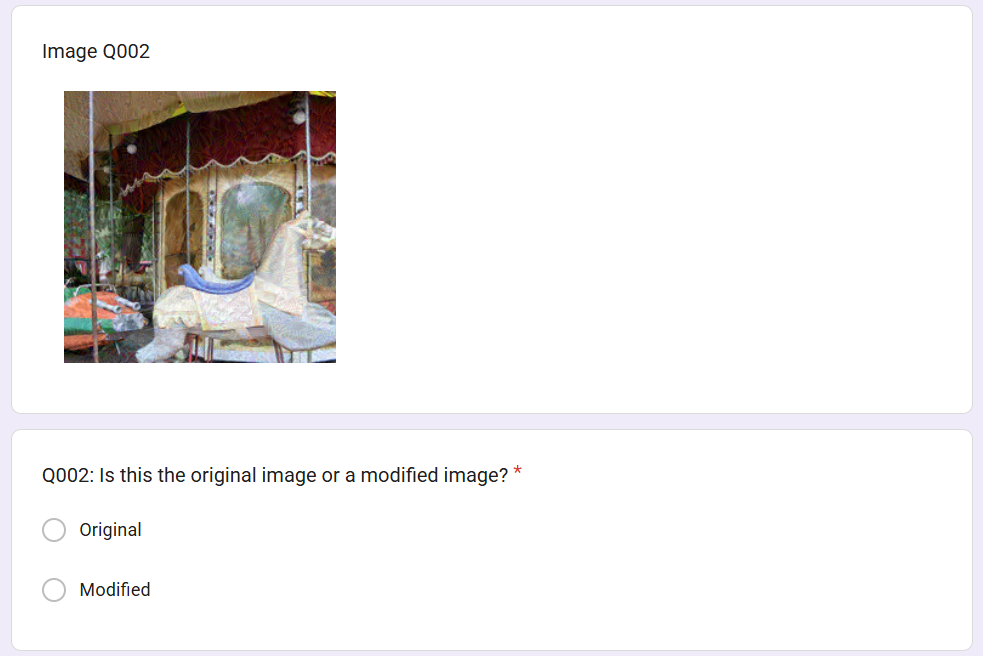}
\caption{\textbf{User study interface.} Participants view a single image and indicate whether it is original or modified.}
\label{fig:user_study_ui}
\end{figure}

We assess perceptual detectability via a binary user study: participants are shown a single image and indicate whether it is \emph{original} or \emph{modified} (Fig.~\ref{fig:user_study_ui}).
We evaluate 50 clean images and 50 adversarial images per method (P2FA, GI-FGSM, DiffAttack, and \ourmethod{}), using 8 participants.
We report the fooling rate (fraction judged as \emph{original}); results are summarized in Fig.~\ref{fig:asr_vs_fooling}.

\section{Additional Qualitative and Spectral Visualizations}
\label{sec:supp_extra_vis}
\begin{figure}[h]
\centering
\includegraphics[width=\linewidth]{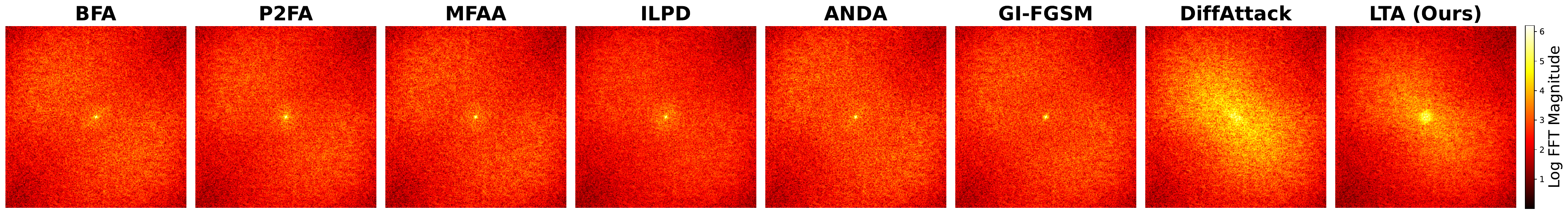}
\caption{\textbf{Additional 2D Fourier spectra.} Log-magnitude of the 2D FFT of $\delta=x_{\text{adv}}-x$ (DC centered) for all methods. Concentration near the center indicates low-frequency perturbations.}
\label{fig:supp_fft2d}
\end{figure}

\begin{figure}[h]
\centering
\includegraphics[width=\linewidth]{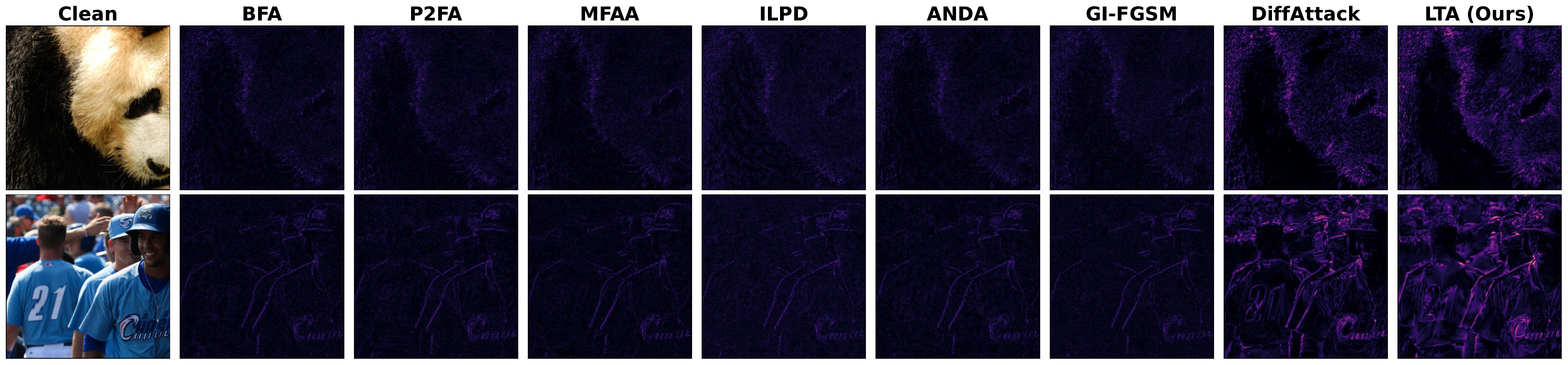}
\caption{\textbf{Additional perturbation magnitude visualizations.} Per-pixel perturbation magnitude $\|\delta(p)\|_2$ for all methods, where $\delta=x_{\text{adv}}-x$. Values are normalized per-image for visualization.}
\label{fig:supp_perturb_mag}
\end{figure}

We provide extended qualitative perturbation magnitude maps (Fig.~\ref{fig:supp_perturb_mag}) and the corresponding 2D FFT spectra (Fig.~\ref{fig:supp_fft2d}) for all methods, computed using the same procedures as in Sec.~\ref{sec:freq_analysis}. These figures are included for completeness and to illustrate that the observed trends are consistent across methods.

\section{Attacks Runtime}
\label{sec:attacks_runtime}
We benchmark per-image runtime on a random subset of 100 images using ResNet-50 as the surrogate model.
All attacks are evaluated with batch size $1$ on a Tesla V100-SXM2-32GB GPU.
Table~\ref{tab:runtime} reports the mean and standard deviation of wall-clock time per image.
Among the baselines, GI-FGSM is the fastest (0.53s), while optimization-heavy methods such as ILPD and P2FA incur substantially higher cost.
Our latent optimization method is slower due to iterative latent updates and VAE decoding, and diffusion-based attacks are the most expensive.

\begin{table}[t]
\centering
\caption{\textbf{Runtime per image} (100 images, surrogate = ResNet-50). We report mean $\pm$ std wall-clock time (seconds) per image.}
\label{tab:runtime}
\small
\setlength{\tabcolsep}{2pt}
\renewcommand{\arraystretch}{1.05}
\begin{tabular}{lc}
\toprule
\textbf{Attack} & \textbf{Time (s/image)} \\
\midrule
GI-FGSM     & $\textbf{0.53} \pm \textbf{0.01}$ \\
ANDA        & $1.59 \pm 0.03$ \\
MFAA        & $1.80 \pm 0.01$ \\
BFA         & $2.15 \pm 0.09$ \\
ILPD        & $4.89 \pm 0.16$ \\
P2FA        & $22.71 \pm 0.71$ \\
DiffAttack  & $68.42 \pm 0.17$ \\
\ourmethod{} (Ours)  & $38.14 \pm 0.57$ \\
\bottomrule
\end{tabular}
\end{table}

\end{document}